\documentclass[11pt]{article}

\usepackage[preprint]{acl}

 \usepackage{microtype}

\usepackage[english,bidi=default,shorthands=off]{babel} 
\babelfont{rm}{TeXGyreTermesX} 

\babelprovide[import]{hindi}
\babelfont[*devanagari]{rm}[
    Path = ./fonts/,
    Extension = .ttf,
    UprightFont = Lohit-Devanagari,
    Script = Devanagari
]{Lohit-Devanagari}

\babelprovide[import]{arabic}
\babelfont[*arabic]{rm}[
    Path = ./fonts/,
    Extension = .ttf,
    UprightFont = NotoSansArabic-Regular,
    Script = Arabic
]{NotoSansArabic-Regular}

\babelprovide[import]{russian}
\babelfont[russian]{rm}[
    Path = ./fonts/,
    Extension = .ttf,
    UprightFont = CharisSIL-Regular,
    BoldFont = CharisSIL-Bold
]{CharisSIL-Regular}

\babelprovide[import]{japanese}
\babelfont[japanese]{rm}[
    Extension = .otf,
    UprightFont = HaranoAjiMincho-Regular
]{HaranoAjiMincho-Regular}

\babelprovide[import]{chinese}
\babelfont[chinese]{rm}[
    Extension = .otf,
    UprightFont = FandolSong-Regular
]{FandolSong-Regular}

\usepackage{booktabs}
\usepackage{adjustbox}
\usepackage{url}
\usepackage{todonotes}
\usepackage{xspace}
\usepackage{subcaption}
\usepackage{amsmath}
\usepackage{float}
\usepackage{placeins}

\newcommand{\rparagraph}[1]{\vspace{1.4mm}\noindent\textbf{#1.}}
\newcommand{\rrparagraph}[1]{\vspace{1.2mm}\noindent\textit{#1.}}
\newcommand{\sparagraph}[1]{\vspace{0.0mm}\noindent\textbf{#1.}}

\newcommand{\mono}{Mono\xspace}
\newcommand{\multi}{Multi\xspace}
\newcommand{\moortho}{MonoNat\xspace}
\newcommand{\mouroman}{MonoUroman\xspace}
\newcommand{\mouconv}{MonoUconv\xspace}
\newcommand{\muortho}{MultiNat\xspace}
\newcommand{\muroman}{MultiRom\xspace}
\newcommand{\tmoortho}{MonoNat\xspace}
\newcommand{\tmouroman}{MonoUroman}
\newcommand{\tmouconv}{MonoUconv}
\newcommand{\tmuortho}{MultiNat}
\newcommand{\tmuroman}{MultiRom}
\newcommand{\tokhp}{Tok.}
\newcommand{\vocab}{Vocab}
\newcommand{\token}{Token Avg.}
\newcommand{\sequence}{Seq. Avg.}
\newcommand{\lm}{LM\xspace}
\newcommand{\lms}{LMs\xspace}

\title{One Script Instead of Hundreds? \\ On Pretraining Romanized Encoder Language Models}

\author{Benedikt Ebing\thanks{Equal contribution.}, Lennart Keller\footnotemark[1] \and Goran Glavaš \\
          Center for Artificial Intelligence and Data Science \\ University of Würzburg \\ \texttt{\{firstname\}.\{lastname\}@uni-wuerzburg.de}}

\begin{document}
\maketitle
\begin{abstract}

Exposing latent lexical overlap, script romanization has emerged as an effective strategy for improving cross-lingual transfer (XLT) in multilingual language models (mLMs). Most prior work, however, focused on setups that favor romanization the most: \textbf{(1)} transfer from high-resource Latin-script to low-resource non-Latin-script languages and/or \textbf{(2)} between genealogically closely related languages with different scripts. It thus remains unclear whether romanization is a good representation choice for \textit{pretraining} general-purpose mLMs, or, more precisely, if information loss associated with romanization harms performance for high-resource languages. 
We address this gap by pretraining encoder \lms from scratch on both romanized and original texts for six typologically diverse high-resource languages, investigating two potential sources of degradation: \textbf{(i)} loss of script-specific information and \textbf{(ii)} negative cross-lingual interference from increased vocabulary overlap.
Using two romanizers with different fidelity profiles, we observe negligible performance loss for languages with segmental scripts, whereas languages with morphosyllabic scripts (Chinese and Japanese) suffer degradation that higher-fidelity romanization mitigates but cannot fully recover. Importantly, comparing monolingual LMs with their mLM counterpart, we find no evidence that increased subword overlap induces negative interference. 
We further show that romanization improves encoding efficiency (i.e., fertility) for segmental scripts at a negligible performance cost.
\end{abstract}


\section{Motivation and Background}
Encoder LMs still outperform their comparable-size (or even larger) decoder-only counterparts on many natural language understanding (NLU) tasks, like classification, clustering, or retrieval \cite{weller2025seqvsseqopen,gisserotboukhlef2025pretrainencodersmaskedlanguage}; and training both monolingual \cite{portes2023MosaicBERT,breton2025neobertnextgenerationbert,ehrmanntraut2025moderngbert,warner-etal-2025-smarter} and multilingual encoder \lms has recently regained traction \cite{marone2025mmbertmodernmultilingualencoder, boizard2025eurobertscalingmultilingualencoders}. 

Multilingual LMs (mLMs) have long been used as primary vehicles for downstream cross-lingual transfer (XLT) \citep{dufter-schutze-2020-identifying,muller-etal-2021-first}: fine-tuned on task data in a source language, they perform the same task in a target language with few \cite[\textit{few-shot XLT}]{xu-murray-2022-por,schmidt-etal-2022-dont} or no labeled task examples \cite[\textit{zero-shot XLT}]{pires-etal-2019-multilingual,lauscher-etal-2020-zero,schmidt-etal-2023-free}.   

More recently, \textit{romanization}---converting non-Latin scripts to the Latin alphabet---has emerged as a tool to break the script barrier in multilingual language learning \cite{ma-etal-2025-exploring-role, xhelili-etal-2024-breaking, j-etal-2024-romansetu}.
Mapping diverse scripts to a common symbolic representation, romanization exposes latent lexical overlap and thus directly facilitates XLT with mLMs \cite{amrhein-sennrich-2020-romanization, dabre-etal-2022-indicbart, moosa-etal-2023-transliteration}.
Romanization is particularly suited for encoder \lms, as we can, unlike in text generation with decoders, avoid the non-trivial additional step of re-mapping the romanized text back to the native script.
Still, being non-injective, romanization is inherently lossy: multiple native-script symbols may map to the same (sequences of) Latin characters. This creates a trade-off between enhanced cross-lingual transfer through script unification and the potential loss of script-specific information.

Prior research on romanization in \lms shows promising results, but studies have predominantly focused on XLT and two scenarios that arguably favor romanization the most: \textbf{(1)} transfer within a language group with known (lexical) similarities, such as genealogically related languages with different scripts (e.g., languages of the Indo-Aryan family) \cite{khemchandani-etal-2021-exploiting, dhamecha-etal-2021-role, liu-etal-2025-transliterations}
, or \textbf{(2)} transfer from a high-resource Latin-script language (usually English) to the text-lean languages from the long tail with non-Latin scripts \cite{muller-etal-2021-unseen, purkayastha-etal-2023-romanization, liu-etal-2024-translico, liu-etal-2025-transmi}.
These are precisely the setups in which the benefit of unlocking lexical overlap outweighs the drawback of information loss \textit{the most}. What remains unclear, however, is how romanization affects the high-resource languages, for which training data is abundant and thus \textbf{(a)} the benefits of exposing lexical links to other languages much smaller and \textbf{(b)} total information loss much larger, as it scales with the training data.     
This gap is critical for assessing whether romanization is a good representation choice
for pretraining \textit{general-purpose} mLMs, which need to exhibit strong in-language performance for high-resource languages as much as robust XLT performance.

\rparagraph{Focused Contribution} In this controlled study, we address this gap by quantifying the downstream performance gaps for high-resource languages between encoder LMs pretrained on romanized vs. native-script data. 
We (pre)train from scratch both monolingual \lms, to isolate the impact of the loss of script-specific information, and corresponding mLMs, to assess whether increased vocabulary overlap leads to negative interference.
%
%
We experiment with six typologically diverse high-resource languages that span six writing systems, and investigate two romanization variants, produced by two popular romanization tools with different fidelity characteristics. Results on five established downstream tasks reveal that romanized \lms show \textit{small to negligible performance losses} compared to native script counterparts, in most cases within one standard deviation. While the losses are more pronounced for Chinese and Japanese, two of our experimental languages with morphosyllabic scripts, we show that higher-fidelity romanization can partially mitigate them. Finally, we analyze the trade-off between information loss and tokenizer's fertility gain: we find that romanized models yield substantially better fertility, while performing only marginally worse than their native-script counterparts.
\section{Romanization Approaches}
Most existing work pre-selects the romanization tool. In contrast, to isolate the downstream effects of the romanizer choice, we compare two widely-used romanizers, which give rise to a trade-off between romanization fidelity and script coverage: 

\rparagraph{URoman}
\texttt{URoman} \cite{hermjakob-etal-2018-box} aims to maximize script coverage by combining
(i) parsing character descriptions from the Unicode codepoint database to derive phonetic Latin correspondences, with
(ii) manual corrections where this heuristic fails and (iii) established script-specific mappings for scripts where Unicode descriptions do not provide reliable cues for Latin equivalents (e.g., Chinese characters).
As a result, \texttt{URoman} offers a near-universal, ASCII-constrained romanization.

\rparagraph{UConv}
\texttt{UConv} bundles standardized romanization schemes, incorporating both transliteration and transcription-focused approaches. While its script coverage is narrower than that of \texttt{URoman}, its adherence to standards involves the use of diacritics, thereby enhancing romanization fidelity by increasing phonemic and orthographic distinctions.

We provide further details and romanization examples in Appendix \ref{app:pretraining}.


\begin{table*}[t!]
\setlength{\tabcolsep}{4.7pt}
\small
\centering
\begin{tabular}{lccccccccc}
\toprule
 & arb\_Arab & cmn\_Hani & hin\_Deva & jpn\_Jpan & rus\_Cyrl & vie\_Latn & \sequence & \token & Avg. \\
\midrule
\tmoortho & 80.3$_{\pm0.6}$ & 78.4$_{\pm0.7}$ & 81.0$_{\pm1.6}$ & 74.9$_{\pm1.2}$ & 81.8$_{\pm0.6}$ & 82.0$_{\pm0.7}$ & 82.2$_{\pm1.0}$ & 76.6$_{\pm1.0}$ & 79.7$_{\pm1.0}$ \\
\tmouroman & 80.7$_{\pm0.8}$ & 74.1$_{\pm1.0}$ & 80.7$_{\pm0.8}$ & 71.7$_{\pm1.1}$ & 81.6$_{\pm0.6}$ & 81.9$_{\pm0.7}$ & 81.5$_{\pm0.8}$ & 74.7$_{\pm0.8}$ & 78.4$_{\pm0.8}$ \\
\tmouconv & 80.3$_{\pm0.8}$ & 76.3$_{\pm1.0}$ & 80.9$_{\pm0.5}$ & 73.6$_{\pm0.9}$ & 82.0$_{\pm0.5}$ & 82.0$_{\pm0.7}$ & 81.9$_{\pm0.8}$ & 75.8$_{\pm0.8}$ & 79.2$_{\pm0.8}$ \\ \midrule
\tmuortho & 77.5$_{\pm1.2}$ & 76.5$_{\pm0.6}$ & 77.6$_{\pm1.6}$ & 72.0$_{\pm1.3}$ & 78.8$_{\pm1.2}$ & 79.3$_{\pm1.1}$ & 79.3$_{\pm1.2}$ & 74.0$_{\pm1.2}$ & 76.9$_{\pm1.2}$ \\
\tmuroman & 77.3$_{\pm1.0}$ & 74.5$_{\pm1.1}$ & 76.5$_{\pm1.0}$ & 69.3$_{\pm0.9}$ & 78.2$_{\pm2.0}$ & 79.0$_{\pm1.4}$ & 78.6$_{\pm1.3}$ & 72.3$_{\pm1.3}$ & 75.8$_{\pm1.3}$ \\
\bottomrule
\end{tabular}
\caption{In-language performance averaged across five tasks with standard deviation ($\pm$). \textit{\sequence\xspace} and \textit{\token} columns aggregate sequence and token classification results.}
\label{tab:results_main}
\vspace{-1em}
\end{table*}

\section{Experiments}
We isolate the downstream effects of romanization for six typologically diverse high-resource languages with different writing systems: Arabic, Chinese, Hindi, Japanese, Russian, and Vietnamese.\footnote{Vietnamese uses Latin script but encodes tonal information through diacritics, which \texttt{URoman} removes.}
Compared to native scripts, romanization introduces two possible causes of degradation: \textbf{(i)} loss of script-specific information and \textbf{(ii)} cross-lingual interference from increased subword overlap.
Note that the nature of the effect of larger vocabulary overlap likely depends on the  \textit{resourceness} of involved languages: while it drives the positive XLT from high- to low-resource languages, it may cause negative interference between high-resource languages, where semantically divergent yet frequently occurring tokens are forced into shared representations. We train the following \lm variants.

\rrparagraph{Monolingual} We train monolingual encoders from scratch to isolate the effect of the script-specific information loss, e.g., the loss of tonal information.
We obtain two romanized \lms for each language: \textbf{\mouroman} and \textbf{\mouconv}, trained on corpora romanized with \texttt{URoman} and \texttt{UConv}, respectively (for each, we also train a dedicated tokenizer).  
For a fair comparison, we train a corresponding model on the original, native-script data of each language: \textbf{\moortho}.
Securing \textit{information parity} in pretraining is key here: for each language, all models are exposed to exactly the same documents and pretrained with the same, fixed computational budget.
We report token counts in Appendix \ref{app:pretraining}.

\rrparagraph{Multilingual} To test if romanization's increase in lexical overlap leads to negative cross-lingual interference, we train multilingual \lms.  
The results of monolingual models (see \S\ref{ssec:results}) informed the choice of romanization tool for each language:
we train \textbf{\muroman} on mixed romanized data, concatenating \texttt{URoman} Arabic, Hindi, and Russian data with \texttt{UConv} Chinese, Japanese, and Vietnamese data.
\xspace We compare \textbf{\muroman} against \textbf{\muortho}, trained on the concatenated original, native-script corpora.



\rparagraph{Model Training}
We sample pretraining data from the Fineweb-2 corpus \cite{penedo2025fineweb2pipelinescale}, containing cleaned, deduplicated, and filtered CommonCrawl texts. 
We train all \lms with the ModernBERT architecture \cite{warner-etal-2025-smarter} in \texttt{Base} size (149M parameters).
We train BPE tokenizers using the \texttt{sentencepiece} library \cite{kudo-richardson-2018-sentencepiece} on 100k randomly sampled documents from the pretraining corpus, with a vocabulary size of 50 048 (unless stated otherwise). We provide further pretraining details in Appendix \ref{app:pretraining}.

\rrparagraph{Fine-Tuning}
We fine-tune for (and evaluate on) five standard tasks, mixing sequence classification: natural language inference (XNLI) \cite{conneau-etal-2018-xnli}, topic classification (SIB200) \cite{adelani-etal-2024-sib}, and intent classification (MASSIVE) \cite{fitzgerald-etal-2023-massive, bastianelli-etal-2020-slurp}; with token classification: named entity recognition (WikiAnn) \cite{pan-etal-2017-cross} and slot filling (MASSIVE) \cite{fitzgerald-etal-2023-massive}. For each experiment, we execute five fine-tuning runs with distinct seeds and report the mean performance. We provide further fine-tuning details in Appendix \ref{app:finetuning}.

\begin{figure*}[ht]
     \centering
     \begin{subfigure}[b]{0.49\textwidth}
         \centering
         \includegraphics[width=0.95\textwidth]{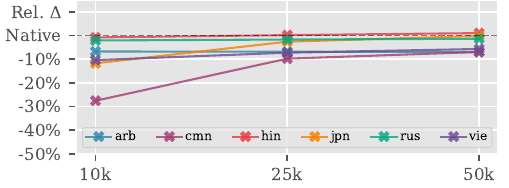}
         \caption{}
         \label{fig:fertility_vs_vocab}
     \end{subfigure}
     \hfill
     \begin{subfigure}[b]{0.49\textwidth}
         \centering
         \includegraphics[width=0.95\textwidth]{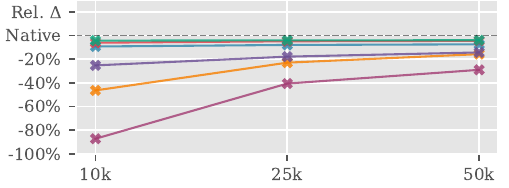}
         \caption{}
         \label{fig:loss_vs_vocab}
     \end{subfigure}
     \vspace{-0.8em}
    \caption{Fertility vs. token merging for monolingual models with \texttt{URoman}: \textbf{(a)} relative change in fertility (lower is better) for \mouroman compared to \moortho; \textbf{(b)} relative change in vocabulary size (higher is better) of \moortho when romanizing its vocabulary with \texttt{URoman} (i.e., subwords conflated due to same romanization).}
    \label{fig:efficiency_loss}
    \vspace{-1em}
    \end{figure*}

\begin{figure}[ht]
\centering
\includegraphics[width=\columnwidth]{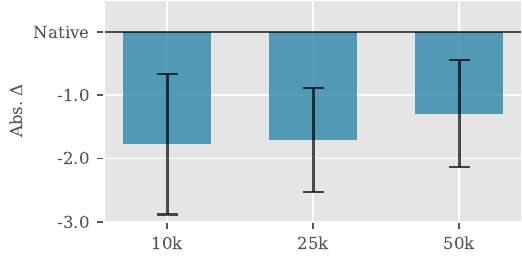}
\caption{Absolute performance difference (avg. across all six languages and five tasks) between \moortho and \mouroman for different vocabulary sizes.}
\label{fig:perf_vs_vocab}
\vspace{-1.5em}
\end{figure}

\section{Results and Analyses}
\label{ssec:results}
\sparagraph{Monolingual Models} Table \ref{tab:results_main} outlines our main results. On average, the information loss introduced by romanization has a limited downstream effect: compared to the native-script models (\moortho), \mouroman counterparts display a marginal average performance drop (-1.3\%), whereas \mouconv variants perform essentially on par (-0.5\%, less than one standard deviation). Our results reveal two groups of writing systems. For languages with largely segmental scripts (Arabic, Hindi, Russian, and Vietnamese), our romanized models perform on par with \moortho for both romanizers. Interestingly, we do not see a notable performance drop for Vietnamese for \mouroman, despite \texttt{URoman} removing suprasegmental tonal information. These results are encouraging, as they suggest that romanization is a safe choice for segmental scripts (i.e., Alphabets, Abjads, and Abugidas), which account for the largest number of world languages, by far.  

In the second group, we find Japanese and Chinese, two writing systems with substantial morphosyllabic components: here, romanization introduces larger downstream losses, as well as more pronounced differences between the two types of romanization: \mouconv outperforms \mouroman by 2.2\% for Chinese (1.9\% for Japanese), but falls behind \moortho by 2.1\% for Chinese (1.3\% for Japanese).
Here, the extent of performance loss seems also dependent on the type of task, with romanized models trailing by a wider margin on token classification tasks. 
%
This aligns with \newcite{schmidt-etal-2022-slicer}, who show that token classification models derive useful features from tokens themselves rather than from the context---an effect that is amplified in WikiAnn, where contexts are often very short.
Furthermore, Chinese and Japanese are typically evaluated at the character level, where romanization introduces far more ambiguity than for longer character sequences.
For sequence classification, however, we observe that both types of romanization perform comparably to \moortho: we attribute this to the fact that longer sequences provide sufficient contextual information to resolve token-level ambiguities introduced by romanization.

\rparagraph{Multilingual Models}
The trends observed for the monolingual models largely hold: \textbf{(i)} \muroman performs within one standard deviation of \muortho, with on par performance for \textbf{(ii)} Arabic, Hindi, Russian, and Vietnamese; \textbf{(iii)} gaps between \muortho and \muroman are consistently smaller on sequence classification tasks than on token classification tasks.
In sum, we observe no evidence that increased subword overlap, introduced by romanization, leads to negative cross-lingual interference in our multilingual models.

\rparagraph{Tokenizer Analysis}
%
Looking through the lens of tokenization, romanization entails a trade-off too.
On the one hand, it reduces character diversity, which allows the tokenizer to learn longer subwords, thereby yielding lower fertility (i.e., fewer tokens for the same text). 
On the other hand, information loss incurred by romanization may adversely affect the model's downstream performance by increasing ambiguity. 
To approximate the information loss of romanization, we measure \textit{token collapse}, the number of tokens in the vocabulary of \moortho that share the romanized form with some other token(s). 
%
Figure \ref{fig:fertility_vs_vocab} illustrates this trade-off between fertility and ambiguity for \mouroman across vocabulary sizes: for larger vocabularies, romanized models suffer a smaller information loss w.r.t. the corresponding native-script model, but also approach the fertility of the native-script model; conversely, smaller vocabularies imply lower fertility but also larger information loss. 
For Arabic, Hindi, and Russian, both  fertility and information loss are relatively stable across vocabulary sizes; for Chinese and Japanese, however,  we see significantly larger information loss and fertility gains with vocabulary reduction; Vietnamese falls in between the two groups. 
These patterns largely hold for multilingual models too (see Appendix \ref{app:information_loss}), albeit with script-specific fertility differences. 

Crucially, as illustrated in Figure \ref{fig:perf_vs_vocab}, the downstream performance seems to be quite robust to vocabulary reduction (i.e., insensitive to increased information loss):  even the models with a vocabulary of 10K tokens exhibit only marginal performance degradation compared to their native-script counterparts, despite substantial token collapse due to romanization (see Figure \ref{fig:fertility_vs_vocab}).
Arabic and Vietnamese illustrate this most clearly: both achieve substantially lower fertility ($-7.0\%$ and $-5.7\%$, respectively), while matching \moortho on downstream tasks ($+0.4\%$ and $-0.1\%$). 
This suggests that for segmental scripts, romanization offers a favorable trade-off: improved encoding efficiency at a negligible cost to task performance.
\section{Conclusion}
We studied the impact of romanization on encoder LM performance across six typologically diverse languages under controlled pretraining conditions.
We find that romanization incurs negligible performance loss for segmental scripts while offering substantial fertility gains---a favorable trade-off for efficient multilingual LM pretraining.
For morphosyllabic scripts (Chinese, Japanese), higher-fidelity romanizers mitigate but do not fully eliminate the performance gap, indicating that such writing systems require special attention.
Importantly, we find that increased subword overlap due to romanization does not lead to negative cross-lingual interference. Combined with known benefits for low-resource XLT, this warrants pretraining of massively multilingual \textit{romanized} LMs as they can improve cross-lingual sharing without sacrificing monolingual performance for high-resource languages.
\section{Limitations}

\rparagraph{Language and Script Coverage}
Given our focus on high-resource languages and computational budget constraints, we investigate only six of the world's approximately 293 writing systems.
While the six corresponding languages greatly vary in terms of linguistic typology and account for a substantial portion of the world's population, many script families remain unexplored.
Furthermore, for each writing system, we examine only a single (arguably most representative) language.
We thus do not study the potentially interesting variation in how romanization affects languages of the same writing system; a comparison between typologically different and genealogically unrelated languages that share the same script and are ideally of ``similar \textit{resourceness}'' (e.g., Persian or Urdu vs. Arabic for the Arabic script, or Kazakh vs. Mongolian for the Cyrillic script).

\rrparagraph{Scale}
Our models are trained at the \texttt{Base} size with fewer pretraining tokens than state-of-the-art multilingual models \cite{marone2025mmbertmodernmultilingualencoder, boizard2025eurobertscalingmultilingualencoders, conneau-etal-2020-unsupervised}.
While this setup allows for controlled comparisons within our computational budget, scaling behavior at larger model sizes and with more training data may, in principle, differ. However, we think it is unlikely that training at larger scales would fundamentally alter our findings, as our scale already suffices to expose nuanced differences in romanization effects (e.g., for segmental vs. morphosyllabic scripts; or for \texttt{Uroman} vs. \texttt{UConv} romanization of morphosyllabic scripts).
\section{Acknowledgements}
Pretraining experiments were conducted using computing resources provided by WestAI on the supercomputer JURECA \cite{thörnig2021jureca} at Jülich Supercomputing Centre (JSC).

Fine-tuning experiments were performed using Julia 2.
Julia 2 was funded as DFG project as "Forschungsgroßgerät nach Art 91b GG" under INST 93/1145-1 FUGG.

\bibliography{custom}

\appendix

\section{Pretraining}
\label{app:pretraining}

\begin{table*}[h]
\setlength{\tabcolsep}{4.4pt}
\centering
\begin{tabular}{lcccccccc}
\toprule
 & \tokhp & Vocab & arb\_Arab & cmn\_Hani & hin\_Deva & jpn\_Jpan & rus\_Cyrl & vie\_Latn \\
\midrule
\tmoortho & no-ws & 10k & 24.71 & 21.61 & 12.49 & 17.15 & 18.27 & 13.92 \\
\tmoortho & ws & 10k & 24.81 & 21.61 & 13.69 & 17.15 & 18.80 & 16.44 \\
\tmoortho & no-ws & 25k & 20.89 & 15.69 & 10.33 & 13.50 & 15.13 & 11.77 \\
\tmoortho & ws & 25k & 21.17 & 15.69 & 12.12 & 13.50 & 15.85 & 15.75 \\
\tmoortho & no-ws & 50k & 18.65 & 14.18 & 9.15 & 11.96 & 13.28 & 10.58 \\
\tmoortho & ws & 50k & 19.24 & 14.17 & 11.43 & 11.96 & 14.17 & 15.19 \\
\tmouroman & no-ws & 10k & 22.96 & 15.77 & 12.34 & 15.13 & 17.85 & 12.45 \\
\tmouroman & ws & 10k & 23.46 & 15.96 & 13.52 & 15.42 & 18.38 & 15.70 \\
\tmouroman & no-ws & 25k & 19.41 & 14.09 & 10.30 & 13.11 & 14.83 & 10.89 \\
\tmouroman & ws & 25k & 20.27 & 14.28 & 12.04 & 13.38 & 15.54 & 15.23 \\
\tmouroman & no-ws & 50k & 17.32 & 13.09 & 9.20 & 11.90 & 13.07 & 9.95 \\
\tmouroman & seg\&ws & 50k & - & 19.21 & - & 20.92 & - & - \\
\tmouroman & ws & 50k & 18.57 & 13.27 & 11.38 & 12.14 & 13.94 & 14.99 \\
\tmouconv & no-ws & 50k & 18.60 & 13.43 & 9.37 & 12.00 & 13.12 & 10.58 \\
\tmuortho & no-ws & 50k & 25.70 & 16.81 & 13.12 & 15.59 & 18.88 & 14.00 \\
\tmuroman & no-ws & 50k & 22.79 & 16.25 & 12.15 & 15.59 & 17.41 & 13.51 \\
\bottomrule
\end{tabular}
\caption{Pretraining token counts (in billions) for the romanized and native script models.}
\label{tab:token_counts}
\end{table*}

\begin{table}[h]
\centering
\scriptsize	
\begin{tabular}{@{}p{1.0\columnwidth}@{}}
\toprule
\textbf{Arabic} \\
\foreignlanguage{arabic}{الناس يولدون أحرارًا ومتساوين.} \\
\texttt{UR:} alnas ywldwn ahrara wmtsawyn. \\
\texttt{UC:} ạlnạs ywldwn ạ̉ḥrạraⁿạ wmtsạwyn. \\
\midrule
\textbf{Devanagari} \\
\foreignlanguage{hindi}{मनुष्य जन्म से स्वतंत्र और समान होते हैं।} \\
\texttt{UR:} manussya janma se svatamtra aur samaan hote haim. \\
\texttt{UC:} manuṣya janma sē svatantra aura samāna hōtē hai\.{m}. \\
\midrule
\textbf{Japanese} \\
\foreignlanguage{japanese}{すべての人は、生まれながら自由で平等である。} \\
\texttt{UR:} subetenorenha, shengmarenagaraziyoudepingdengdearu. \\
\texttt{UC:} subeteno rénha, shēngmarenagara zì yóude píng děngdearu. \\
\midrule
\textbf{Chinese} \\
\foreignlanguage{chinese}{人人生而自由平等。} \\
\texttt{UR:} renrenshengerziyoupingdeng. \\
\texttt{UC:} rén rén shēng ér zì yóu píng děng. \\
\midrule
\textbf{Cyrillic} \\
\foreignlanguage{russian}{Все люди рождаются свободными и равными.} \\
\texttt{UR:} Vse lyudi rozhdayutsya svobodnymi i ravnymi. \\
\texttt{UC:} Vse lûdi roždaûtsâ svobodnymi i ravnymi. \\
\bottomrule
\end{tabular}
\caption{Romanization of \textit{"All human beings are born free and equal"} using \texttt{URoman} (\texttt{UR}) and \texttt{UConv} (\texttt{UC}).}
\label{table:rom-examples}
\end{table}

\sparagraph{Data}
We source the pretraining data from the Fineweb2 corpus \cite{penedo2025fineweb2pipelinescale} through the Hugging Face library. The dataset is already cleaned, deduplicated, and filtered. We do not apply further preprocessing. For a fair comparison, we expose all models within a language to the exact same information (i.e., the same documents). We randomly sample documents and ensure that the number of training tokens is at most equal to the total number of tokens seen during training (i.e., each model trains at least one epoch on all documents). The resulting number of training tokens per model is shown in Table \ref{tab:token_counts}. Fineweb2 is licensed under ODC-By 1.0.

\rparagraph{Romanization Tools}
To efficiently romanize pretraining-scale data, we use a Rust reimplementation of \texttt{URoman}, which yields a roughly 27x speedup in throughput compared with the original Python implementation. Both versions are licensed under the Apache License Version 2.0.

\texttt{UConv} is part of the International Components for Unicode (ICU) (\href{https://icu.unicode.org}{https://icu.unicode.org}), which is open-source-licensed under the Unicode License.
For \texttt{UConv} romanization, we use the default \texttt{any-to-Latin} preset that dynamically selects a suitable romanization scheme.
The scheme selection of \texttt{UConv} generally follows the recommendations of the Working Group on Romanization Systems of the United Nations Group of Experts on Geographical Names (UNGEGN), and thus utilizes both transcription- and transliteration-focused approaches.
Transcription-focused Romanization generally assumes a reference pronunciation of Latin characters, whereas transliteration aims to create invertible character mappings between the source script and the Latin alphabet. Thus, the Russian name “\foreignlanguage{russian}{\textbf{Ч}айковский}” appears as “\textbf{Ch}aĭkovskiĭ” under the ALA–LC transcription scheme aimed at English speakers, whereas the ISO 9 transliteration scheme for Cyrillic yields “\textbf{Č}ajkovskij”.
In this work, the specific schemes employed via \texttt{UConv} were: ADEGN (Arabic), Pinyin (Chinese), ISO 15919 (Hindi), Hepburn (Japanese), and ISO 9 (Russian).
Table \ref{table:rom-examples} presents example outputs for both romanizers.

\begin{table}[h]
\setlength{\tabcolsep}{13pt}
\centering
\begin{tabular}{@{}lcc@{}}
\toprule
                                  & \mono & \multi \\ \midrule
Training Batches                  & 8000                 & 48000                 \\
Max Sequence Lenght               & \multicolumn{2}{c}{1024}                      \\ \midrule
Batch Size                        & \multicolumn{2}{c}{4608}                     \\
Micro Batch Size                  & \multicolumn{2}{c}{96}                       \\ \midrule
Learning Rate                     & \multicolumn{2}{c}{1e-3}                     \\
\hspace{1em} Schedule         & \multicolumn{2}{c}{Linear}                   \\
\hspace{1em} Warmup & \multicolumn{2}{c}{6\%}                     \\
\hspace{1em} Decay & \multicolumn{2}{c}{94\%}                    \\
Weight Decay                      & \multicolumn{2}{c}{1e-5}                     \\ \midrule
Model Initialization              & \multicolumn{2}{c}{Megatron}                 \\ \midrule
Dropout (attn out)                & \multicolumn{2}{c}{0.1}                      \\
Dropout (all other layers)        & \multicolumn{2}{c}{0.0}                      \\ \midrule
Optimizer                         & \multicolumn{2}{c}{Stable AdamW}             \\
\hspace{1em}Betas            & \multicolumn{2}{c}{(0.9, 0.98)}              \\
\hspace{1em}Epsilons          & \multicolumn{2}{c}{1e-6}                     \\ \midrule
Training Hardware                 & \multicolumn{2}{c}{4x H100}                  \\ \bottomrule
\end{tabular}
\caption{Training details adopted from \citet{warner-etal-2025-smarter} and \citet{izsak-etal-2021-train}}
\label{tab:training_hp}
\end{table}

\begin{table}[h]
\setlength{\tabcolsep}{2.5pt}
\centering
\begin{tabular}{@{}ll@{}}
\toprule
         & Value(s)              \\ \midrule
Vocabulary             & 10112 | 25088 | 50048 \\
Unused Tokens         & 111 | 87 | 47         \\
Layers                 & 22                    \\
Hidden Size            & 768                   \\
Transformer Block      & Pre-Norm              \\
Activation Function    & GeLU                  \\
Linear Bias            & False                 \\
Attention              & Multi-head            \\
Attention Heads        & 12                    \\
Global Attention       & Every three layers    \\
Local Attention Window & 128                   \\
Intermediate Size      & 1 152                 \\
GLU Expansion          & 2 304                 \\
Normalization          & Layer Norm            \\
Norm Epsilon           & 1e-5                  \\
Norm Bias              & False                 \\
RoPE theta             & 10 000                \\
\bottomrule
\end{tabular}
\caption{Model design adopted from \citet{warner-etal-2025-smarter}}
\label{tab:model_hp}
\end{table}

\rparagraph{Model}
We train ModernBert models \cite{warner-etal-2025-smarter} in their \texttt{Base} variant with 149M parameters. We adopt our implementation from the ModernBert GitHub repository: \href{https://github.com/AnswerDotAI/ModernBERT}{https://github.com/AnswerDotAI/ModernBERT} which is licensed under the Apache License 2.0. The training details are displayed in Table \ref{tab:training_hp}. We train all models with bfloat16 mixed precision and distributed data parallelism. Our training setup is adapted from ModernBert \cite{warner-etal-2025-smarter} and \citet{izsak-etal-2021-train}, who focus on training BERT models under constrained budget (e.g., 24 GPU hours). The model design hyperparameters are displayed in Table \ref{tab:model_hp}. A single pretraining run for a monolingual model required 34 GPU hours (204 GPU hours for a multilingual model). Overall, we estimate the total pretraining time to be 3000 GPU hours.

\rparagraph{Tokenizer}
\setlength{\tabcolsep}{25pt}
Our tokenizers are trained with the sentencepiece library \cite{kudo-richardson-2018-sentencepiece}, which is licensed under Apache-2.0. We provide the used hyperparameters in Table \ref{tab:tokenizer_hp}. The tokenizers for the multilingual models are trained on 100 000 documents per language (i.e., 600 000 in total).
The models in the main body of the paper are trained with the \textit{Split by Whitespace}-flag set to \textit{true}.
For \texttt{UConv}, this hyperparameter choice is required as the romanizer introduces additional whitespaces during transcription. This is particularly relevant for Chinese and Japanese, where whitespaces are inserted around the romanization of almost every character. Therefore, not setting the flag would result in training a romanized model that can merge tokens only up to the level of a Chinese or Japanese character. Uroman, by contrast, does not introduce additional whitespace. Hence, we ablate the impact of this hyperparameter for Uroman models. Furthermore, we ablate pre-segmenting the Chinese and Japanese text prior to tokenizer training.
For pre-segmentation, we use \texttt{pkuseg} \cite{pkuseg-tokenizer} for Chinese and \texttt{Sudachi} \cite{sudachi-tokenizer} for Japanese (licensed under the MIT License and Apache License 2.0, respectively), both with their default configurations.
After training, we convert the tokenizers to the Hugging Face format using the tokenizers library to facilitate integration into the model training pipeline. The library is licensed under Apache License 2.0.

\begin{table}[h]
\centering
\begin{tabular}{@{}ll@{}}
\toprule
          & Value(s)        \\ \midrule
Model Type              & BPE             \\
Vocab Size              & 10k | 25k | 50k \\
Input Sentence Size     & 100 000         \\
Byte Fallback           & True            \\
Normalization           & NFKC            \\
Character Coverage      & 0.9999          \\
Max. Sentence Length    & 4192            \\
Split by Whitespace     & False | True    \\
Split by Unicode Script & False           \\
Split by Number         & True            \\
Split by Digits         & False           \\
Hard Vocab Limit        & True            \\ \bottomrule
\end{tabular}
\caption{Sentencepiece hyperparameter choices}
\label{tab:tokenizer_hp}
\end{table}

\section{Downstream Fine-Tuning}
\label{app:finetuning}
\begin{table}[H]
\small
\setlength{\tabcolsep}{4pt}
\begin{tabular}{@{}lcccc@{}}
\toprule
                & SIB200       & WikiAnn     & XNLI & MASSIVE   \\ \midrule
Lang            & All          & Other (hin) & All  & All       \\
Train. Size     & 700          & 20k (5k)    & 393k & 11.5k     \\
Val. Size       & 99           & 10k (1k)    & 2.5k & 2k        \\
Test Size       & 204          & 10k (1k)    & 5k   & 3k        \\
Metric          & Acc.         & F1          & Acc.  & Acc./F1    \\ \midrule
License         & cc-by-sa-4.0 & -           & -    & cc-by-4.0
\end{tabular}
\caption{Dataset sizes, metrics, and licenses for downstream fine-tuning}
\label{tab:finetuning_data}
\end{table}
We begin downstream fine-tuning from the final pretraining checkpoint. For each task, we train models for 10 epochs (5 epochs for XNLI), with a batch size of 32. We use a linear learning rate schedule with 10\% warmup and a weight decay of 0.01. The used optimizer is AdamW with a learning rate of 2e-5. We pick the best checkpoint based on the in-language validation performance. We employ the Hugging Face Transformers library for our downstream experiments, which is licensed under the Apache 2.0 license. For MASSIVE, we use the official downstream evaluation repository: \href{https://github.com/alexa/massive}{https://github.com/alexa/massive} licensed under Apache License 2.0. All fine-tuning experiments were run on L40 GPUs with 40GB of VRAM. We estimate the total compute to 2000 GPU hours. We comply with the licenses governing the downstream datasets used. Details on the dataset sizes and evaluation metrics for each dataset are provided in Table \ref{tab:finetuning_data}.

\section{Details on Fertility}
We compute the fertility of each tokenizer on the respective test sets from Fineweb2 using the following formula:
\begin{equation*}
    Fertility_{\moortho} = \frac{\#Tokens}{\#Words}
\end{equation*}
, where\xspace $\#Tokens$ refers to the total number of tokens used by the respective tokenizer (e.g., \moortho, \mouroman, \dots) to encode the documents in the test set and $\#Words$ refers to the number of words in the native test set. Using the number of words in the original data test set ensures that the fertility scores are comparable across tokenizers. For Arabic, Hindi, Russian, and Vietnamese, we compute $\#Words$ by splitting on whitespace, whereas for Chinese and Japanese, we use the number of characters as the denominator. Finally, we compute the relative difference in fertility by:
\begin{equation*}
    1 - \frac{Fertility_{\mouroman}}{Fertiltity_{\moortho}}
\end{equation*}

\section{Details on Loss of information}
The loss of information is computed as the difference between the number of tokens remaining after romanizing each entry in the native script tokenizer and the total number of tokens before romanization. We compute the measure by encoding the respective test sets of the Fineweb2 dataset with the native script tokenizers (i.e., \moortho and \muortho). We identify the unique tokens yielding $\#UniqueOrigTokens$. Then, we romanize each unique original token identified in the first step to compute $\#UniqueRomanizedTokens$. Finally, we compute the relative loss in performance by:
\begin{equation*}
   1 - \frac{\#UniqueRomanizedTokens}{\#UniqueOrigTokens} 
\end{equation*}

\begin{table*}[t]
\setlength{\tabcolsep}{2.8pt}
\section{Detailed Results: Main Results and Impact of Tokenization}
\centering
\small
\begin{tabular}{lcccccccccc}
\toprule
 & \tokhp & arb\_Arab & cmn\_Hani & hin\_Deva & jpn\_Jpan & rus\_Cyrl & vie\_Latn & \sequence & \token & Avg. \\
\midrule
\tmoortho & no-ws & 80.3$_{\pm0.6}$ & 78.4$_{\pm0.7}$ & 81.0$_{\pm1.6}$ & 74.9$_{\pm1.2}$ & 81.8$_{\pm0.6}$ & 82.0$_{\pm0.7}$ & 82.2$_{\pm1.0}$ & 76.6$_{\pm1.0}$ & 79.7$_{\pm1.0}$ \\
\tmoortho & ws & 79.8$_{\pm0.9}$ & 78.4$_{\pm1.0}$ & 81.3$_{\pm1.0}$ & 74.6$_{\pm1.7}$ & 81.8$_{\pm0.8}$ & 82.4$_{\pm0.5}$ & 82.0$_{\pm1.0}$ & 76.9$_{\pm1.1}$ & 79.7$_{\pm1.1}$ \\
\tmouroman & no-ws & 80.7$_{\pm0.8}$ & 74.1$_{\pm1.0}$ & 80.7$_{\pm0.8}$ & 71.7$_{\pm1.1}$ & 81.6$_{\pm0.6}$ & 81.9$_{\pm0.7}$ & 81.5$_{\pm0.8}$ & 74.7$_{\pm0.8}$ & 78.4$_{\pm0.8}$ \\
\tmouroman & ws & 80.1$_{\pm1.1}$ & 74.0$_{\pm0.8}$ & 80.8$_{\pm0.8}$ & 72.1$_{\pm0.8}$ & 81.7$_{\pm0.9}$ & 81.3$_{\pm0.8}$ & 81.1$_{\pm0.9}$ & 74.9$_{\pm0.9}$ & 78.3$_{\pm0.9}$ \\
\tmouroman & seg\&ws & - & 74.6$_{\pm0.8}$ & - & 72.2$_{\pm1.5}$ & - & - & 81.1$_{\pm1.1}$ & 64.1$_{\pm1.2}$ & 73.4$_{\pm1.2}$ \\
\tmouconv & no-ws & 80.3$_{\pm0.8}$ & 76.3$_{\pm1.0}$ & 80.9$_{\pm0.5}$ & 73.6$_{\pm0.9}$ & 82.0$_{\pm0.5}$ & 82.0$_{\pm0.7}$ & 81.9$_{\pm0.8}$ & 75.8$_{\pm0.8}$ & 79.2$_{\pm0.8}$ \\ \midrule
\tmuortho & no-ws & 77.5$_{\pm1.2}$ & 76.5$_{\pm0.6}$ & 77.6$_{\pm1.6}$ & 72.0$_{\pm1.3}$ & 78.8$_{\pm1.2}$ & 79.3$_{\pm1.1}$ & 79.3$_{\pm1.2}$ & 74.0$_{\pm1.2}$ & 76.9$_{\pm1.2}$ \\ 
\tmuroman & no-ws & 77.3$_{\pm1.0}$ & 74.5$_{\pm1.1}$ & 76.5$_{\pm1.0}$ & 69.3$_{\pm0.9}$ & 78.2$_{\pm2.0}$ & 79.0$_{\pm1.4}$ & 78.6$_{\pm1.3}$ & 72.3$_{\pm1.3}$ & 75.8$_{\pm1.3}$ \\
\bottomrule
\end{tabular}
\caption{Main results averaged across our 5 evaluation tasks for monolingual and multilingual romanized models. We ablate different settings for the tokenizer training flag \textit{Split by Whitespace}: \texttt{ws} refers to \textit{Split by Whitespace} set to \textit{true}, \texttt{no-ws} refers to \textit{Split by Whitespace} set to \textit{false}, and \texttt{seg\&ws} refers to pre-segmented data for tokenizer training in combination with \textit{Split by Whitespace} set to \textit{true}.}
\label{tab:appendix_main}
\end{table*}

\begin{table*}[t]
\setlength{\tabcolsep}{4pt}
\centering
\begin{tabular}{lcccccccc}
\toprule
 & \tokhp & arb\_Arab & cmn\_Hani & hin\_Deva & jpn\_Jpan & rus\_Cyrl & vie\_Latn & Avg. \\
\midrule
\tmoortho & no-ws & 86.2$_{\pm0.9}$ & 89.4$_{\pm1.0}$ & 85.9$_{\pm3.0}$ & 87.3$_{\pm2.0}$ & 86.8$_{\pm0.8}$ & 87.2$_{\pm0.9}$ & 87.1$_{\pm1.7}$ \\
\tmoortho & ws & 83.0$_{\pm1.2}$ & 88.3$_{\pm1.3}$ & 84.9$_{\pm1.7}$ & 86.6$_{\pm3.3}$ & 85.8$_{\pm1.3}$ & 88.8$_{\pm0.7}$ & 86.2$_{\pm1.8}$ \\
\tmouroman & no-ws & 88.2$_{\pm1.3}$ & 84.5$_{\pm1.7}$ & 84.5$_{\pm1.1}$ & 84.7$_{\pm2.0}$ & 85.8$_{\pm0.7}$ & 89.6$_{\pm1.1}$ & 86.2$_{\pm1.4}$ \\
\tmouroman & ws & 85.3$_{\pm2.2}$ & 84.7$_{\pm1.4}$ & 84.6$_{\pm1.2}$ & 84.7$_{\pm1.3}$ & 85.6$_{\pm1.5}$ & 87.9$_{\pm1.1}$ & 85.5$_{\pm1.5}$ \\
\tmouroman & seg\&ws & - & 87.1$_{\pm1.0}$ & - & 84.8$_{\pm2.8}$ & - & - & 85.9$_{\pm2.1}$ \\
\tmouconv & no-ws & 85.5$_{\pm1.1}$ & 85.7$_{\pm2.1}$ & 85.8$_{\pm0.8}$ & 86.4$_{\pm1.3}$ & 86.8$_{\pm0.5}$ & 87.2$_{\pm0.9}$ & 86.2$_{\pm1.2}$ \\ \midrule
\tmuortho & no-ws & 81.9$_{\pm2.3}$ & 83.0$_{\pm1.0}$ & 78.8$_{\pm3.3}$ & 84.7$_{\pm1.8}$ & 78.5$_{\pm2.2}$ & 82.5$_{\pm1.9}$ & 81.6$_{\pm2.2}$ \\
\tmuroman & no-ws & 82.3$_{\pm1.3}$ & 82.6$_{\pm1.7}$ & 76.0$_{\pm1.1}$ & 82.3$_{\pm1.3}$ & 79.3$_{\pm4.1}$ & 81.8$_{\pm2.6}$ & 80.7$_{\pm2.3}$ \\
\bottomrule
\end{tabular}
\caption{SIB200 results for monolingual and multilingual romanized models. We ablate different settings for the tokenizer training flag \textit{Split by Whitespace}: \texttt{ws} refers to \textit{Split by Whitespace} set to \textit{true}, \texttt{no-ws} refers to \textit{Split by Whitespace} set to \textit{false}, and \texttt{seg\&ws} refers to pre-segmented data for tokenizer training in combination with \textit{Split by Whitespace} set to \textit{true}.}
\label{tab:results_sib200}
\end{table*}

\begin{table*}[t]
\setlength{\tabcolsep}{4pt}
\centering
\begin{tabular}{lcccccccc}
\toprule
 & \tokhp & arb\_Arab & cmn\_Hani & hin\_Deva & jpn\_Jpan & rus\_Cyrl & vie\_Latn & Avg. \\
\midrule
\tmoortho & no-ws & 89.7$_{\pm0.1}$ & 75.4$_{\pm0.2}$ & 90.1$_{\pm0.5}$ & 65.9$_{\pm0.2}$ & 87.3$_{\pm0.1}$ & 90.5$_{\pm0.1}$ & 83.2$_{\pm0.2}$ \\
\tmoortho & ws & 89.9$_{\pm0.1}$ & 75.3$_{\pm0.5}$ & 91.3$_{\pm0.5}$ & 65.9$_{\pm0.3}$ & 87.6$_{\pm0.1}$ & 91.0$_{\pm0.2}$ & 83.5$_{\pm0.3}$ \\
\tmouroman & no-ws & 89.6$_{\pm0.2}$ & 68.1$_{\pm0.4}$ & 89.5$_{\pm0.7}$ & 59.7$_{\pm0.1}$ & 87.1$_{\pm0.3}$ & 89.5$_{\pm0.3}$ & 80.5$_{\pm0.4}$ \\
\tmouroman & ws & 89.7$_{\pm0.2}$ & 67.9$_{\pm0.2}$ & 90.4$_{\pm0.4}$ & 59.3$_{\pm0.3}$ & 87.4$_{\pm0.1}$ & 90.2$_{\pm0.1}$ & 80.8$_{\pm0.2}$ \\
\tmouroman & seg\&ws & - & 68.9$_{\pm0.2}$ & - & 60.6$_{\pm0.3}$ & - & - & 64.8$_{\pm0.3}$ \\
\tmouconv & no-ws & 89.3$_{\pm0.2}$ & 71.4$_{\pm0.2}$ & 89.5$_{\pm0.2}$ & 63.4$_{\pm0.3}$ & 87.3$_{\pm0.2}$ & 90.5$_{\pm0.1}$ & 81.9$_{\pm0.2}$ \\ \midrule
\tmuortho & no-ws & 87.6$_{\pm0.2}$ & 73.8$_{\pm0.3}$ & 88.3$_{\pm0.5}$ & 62.6$_{\pm0.6}$ & 85.8$_{\pm0.1}$ & 89.3$_{\pm0.3}$ & 81.2$_{\pm0.4}$ \\
\tmuroman & no-ws & 87.6$_{\pm0.2}$ & 69.3$_{\pm0.5}$ & 87.2$_{\pm1.3}$ & 59.3$_{\pm0.1}$ & 85.2$_{\pm0.2}$ & 89.1$_{\pm0.1}$ & 79.6$_{\pm0.6}$ \\
\bottomrule
\end{tabular}
\caption{Wikiann results for monolingual and multilingual romanized models. We ablate different settings for the tokenizer training flag \textit{Split by Whitespace}: \texttt{ws} refers to \textit{Split by Whitespace} set to \textit{true}, \texttt{no-ws} refers to \textit{Split by Whitespace} set to \textit{false}, and \texttt{seg\&ws} refers to pre-segmented data for tokenizer training in combination with \textit{Split by Whitespace} set to \textit{true}.}
\label{tab:results_wikiann}
\end{table*}

\begin{table*}[t]
\setlength{\tabcolsep}{7pt}
\centering
\begin{tabular}{lccccccc}
\toprule
 & \tokhp & arb\_Arab & cmn\_Hani & hin\_Deva & rus\_Cyrl & vie\_Latn & Avg. \\
\midrule
\tmoortho & no-ws & 74.6$_{\pm0.6}$ & 74.8$_{\pm0.7}$ & 72.0$_{\pm1.5}$ & 75.7$_{\pm1.0}$ & 76.7$_{\pm1.1}$ & 74.8$_{\pm1.1}$ \\
\tmoortho & ws & 74.2$_{\pm1.3}$ & 75.3$_{\pm1.0}$ & 72.8$_{\pm1.0}$ & 75.6$_{\pm0.9}$ & 76.4$_{\pm0.7}$ & 74.9$_{\pm1.0}$ \\
\tmouroman & no-ws & 74.2$_{\pm0.7}$ & 71.2$_{\pm1.2}$ & 73.3$_{\pm0.6}$ & 75.9$_{\pm0.8}$ & 76.3$_{\pm0.9}$ & 74.2$_{\pm0.9}$ \\
\tmouroman & ws & 74.9$_{\pm0.9}$ & 70.7$_{\pm0.8}$ & 72.9$_{\pm0.7}$ & 75.4$_{\pm1.1}$ & 74.5$_{\pm0.7}$ & 73.7$_{\pm0.9}$ \\
\tmouroman & seg\&ws & - & 70.8$_{\pm1.2}$ & - & - & - & 70.8$_{\pm1.2}$ \\
\tmouconv & no-ws & 74.2$_{\pm1.2}$ & 74.7$_{\pm0.8}$ & 73.0$_{\pm0.7}$ & 75.6$_{\pm0.8}$ & 76.7$_{\pm1.1}$ & 74.9$_{\pm0.9}$ \\ \midrule
\tmuortho & no-ws & 74.3$_{\pm0.8}$ & 74.6$_{\pm0.5}$ & 70.0$_{\pm0.7}$ & 74.9$_{\pm0.3}$ & 76.7$_{\pm0.8}$ & 74.1$_{\pm0.6}$ \\
\tmuroman & no-ws & 74.3$_{\pm0.6}$ & 75.2$_{\pm0.7}$ & 70.9$_{\pm0.7}$ & 74.1$_{\pm1.1}$ & 75.3$_{\pm0.8}$ & 73.9$_{\pm0.8}$ \\
\bottomrule
\end{tabular}
\caption{XNLI results for monolingual and multilingual romanized models. We ablate different settings for the tokenizer training flag \textit{Split by Whitespace}: \texttt{ws} refers to \textit{Split by Whitespace} set to \textit{true}, \texttt{no-ws} refers to \textit{Split by Whitespace} set to \textit{false}, and \texttt{seg\&ws} refers to pre-segmented data for tokenizer training in combination with \textit{Split by Whitespace} set to \textit{true}.}
\label{tab:results_xnli}
\end{table*}

\begin{table*}[t]
\setlength{\tabcolsep}{4pt}
\centering
\begin{tabular}{lcccccccc}
\toprule
 & \tokhp & arb\_Arab & cmn\_Hani & hin\_Deva & jpn\_Jpan & rus\_Cyrl & vie\_Latn & Avg. \\
\midrule
\tmoortho & no-ws & 79.8$_{\pm0.5}$ & 83.6$_{\pm0.5}$ & 85.1$_{\pm0.5}$ & 83.2$_{\pm0.5}$ & 84.8$_{\pm0.5}$ & 85.0$_{\pm0.2}$ & 83.6$_{\pm0.5}$ \\
\tmoortho & ws & 79.9$_{\pm0.5}$ & 83.3$_{\pm1.3}$ & 85.1$_{\pm0.5}$ & 83.3$_{\pm0.5}$ & 85.3$_{\pm0.6}$ & 85.3$_{\pm0.3}$ & 83.7$_{\pm0.7}$ \\
\tmouroman & no-ws & 79.4$_{\pm0.6}$ & 81.4$_{\pm0.4}$ & 84.5$_{\pm0.9}$ & 82.1$_{\pm0.3}$ & 85.3$_{\pm0.6}$ & 84.2$_{\pm0.6}$ & 82.8$_{\pm0.6}$ \\
\tmouroman & ws & 79.3$_{\pm0.3}$ & 81.5$_{\pm0.6}$ & 84.6$_{\pm0.6}$ & 82.4$_{\pm0.4}$ & 85.5$_{\pm0.6}$ & 84.4$_{\pm0.8}$ & 83.0$_{\pm0.6}$ \\
\tmouroman & seg\&ws & - & 81.0$_{\pm0.1}$ & - & 82.0$_{\pm0.6}$ & - & - & 81.5$_{\pm0.5}$ \\
\tmouconv & no-ws & 80.2$_{\pm0.6}$ & 82.4$_{\pm0.3}$ & 84.6$_{\pm0.1}$ & 82.7$_{\pm0.5}$ & 85.2$_{\pm0.4}$ & 85.0$_{\pm0.2}$ & 83.3$_{\pm0.4}$ \\ \midrule
\tmuortho & no-ws & 76.2$_{\pm0.8}$ & 82.5$_{\pm0.3}$ & 82.0$_{\pm0.6}$ & 81.3$_{\pm0.7}$ & 83.1$_{\pm0.7}$ & 82.6$_{\pm0.7}$ & 81.3$_{\pm0.7}$ \\
\tmuroman & no-ws & 75.8$_{\pm1.4}$ & 80.9$_{\pm0.4}$ & 81.1$_{\pm0.6}$ & 79.5$_{\pm0.6}$ & 82.6$_{\pm0.6}$ & 82.5$_{\pm0.7}$ & 80.4$_{\pm0.8}$ \\
\bottomrule
\end{tabular}
\caption{MASSIVE Intent results for monolingual and multilingual romanized models. We ablate different settings for the tokenizer training flag \textit{Split by Whitespace}: \texttt{ws} refers to \textit{Split by Whitespace} set to \textit{true}, \texttt{no-ws} refers to \textit{Split by Whitespace} set to \textit{false}, and \texttt{seg\&ws} refers to pre-segmented data for tokenizer training in combination with \textit{Split by Whitespace} set to \textit{true}.}
\label{tab:results_massive_intent}
\end{table*}

\begin{table*}[t]
\setlength{\tabcolsep}{4pt}
\centering
\begin{tabular}{lcccccccc}
\toprule
 & \tokhp & arb\_Arab & cmn\_Hani & hin\_Deva & jpn\_Jpan & rus\_Cyrl & vie\_Latn & Avg. \\
\midrule
\tmoortho & no-ws & 71.3$_{\pm0.6}$ & 68.6$_{\pm0.9}$ & 71.9$_{\pm0.4}$ & 63.3$_{\pm1.4}$ & 74.6$_{\pm0.1}$ & 70.5$_{\pm0.6}$ & 70.0$_{\pm0.8}$ \\
\tmoortho & ws & 72.1$_{\pm0.6}$ & 69.5$_{\pm0.6}$ & 72.4$_{\pm0.6}$ & 62.6$_{\pm0.8}$ & 74.7$_{\pm0.3}$ & 70.5$_{\pm0.2}$ & 70.3$_{\pm0.6}$ \\
\tmouroman & no-ws & 72.2$_{\pm0.8}$ & 65.2$_{\pm0.6}$ & 71.6$_{\pm0.5}$ & 60.4$_{\pm0.8}$ & 74.2$_{\pm0.4}$ & 69.7$_{\pm0.6}$ & 68.9$_{\pm0.6}$ \\
\tmouroman & ws & 71.5$_{\pm0.5}$ & 65.2$_{\pm0.5}$ & 71.5$_{\pm0.6}$ & 62.0$_{\pm0.8}$ & 74.4$_{\pm0.6}$ & 69.6$_{\pm0.9}$ & 69.0$_{\pm0.7}$ \\
\tmouroman & seg\&ws & - & 65.4$_{\pm0.9}$ & - & 61.4$_{\pm0.5}$ & - & - & 63.4$_{\pm0.8}$ \\
\tmouconv & no-ws & 72.2$_{\pm0.7}$ & 67.3$_{\pm0.7}$ & 71.7$_{\pm0.5}$ & 61.8$_{\pm1.0}$ & 74.9$_{\pm0.2}$ & 70.5$_{\pm0.6}$ & 69.7$_{\pm0.7}$ \\ \midrule
\tmuortho & no-ws & 67.8$_{\pm1.0}$ & 68.5$_{\pm0.3}$ & 68.7$_{\pm0.7}$ & 59.3$_{\pm1.7}$ & 71.7$_{\pm1.3}$ & 65.1$_{\pm1.3}$ & 66.8$_{\pm1.1}$ \\
\tmuroman & no-ws & 66.9$_{\pm1.2}$ & 64.5$_{\pm1.4}$ & 67.1$_{\pm1.1}$ & 56.3$_{\pm1.3}$ & 69.6$_{\pm1.1}$ & 66.1$_{\pm1.4}$ & 65.1$_{\pm1.2}$ \\
\bottomrule
\end{tabular}
\caption{MASSIVE Slot Filling results for monolingual and multilingual romanized models. We ablate different settings for the tokenizer training flag \textit{Split by Whitespace}: \texttt{ws} refers to \textit{Split by Whitespace} set to \textit{true}, \texttt{no-ws} refers to \textit{Split by Whitespace} set to \textit{false}, and \texttt{seg\&ws} refers to pre-segmented data for tokenizer training in combination with \textit{Split by Whitespace} set to \textit{true}.}
\label{tab:results_massive_slot}
\end{table*}

\begin{figure*}[h]
\section{Detailed Results: Fertility vs. Information Loss}
\label{app:information_loss}
     \centering
     \begin{subfigure}[b]{0.49\textwidth}
         \centering
         \includegraphics{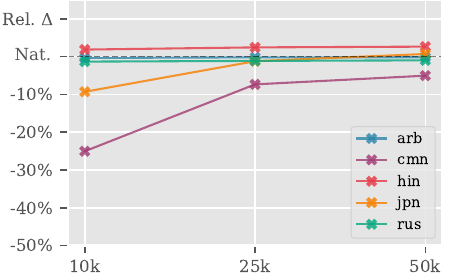}
         \caption{}
         \label{fig:uconv_fertility_vs_vocab}
     \end{subfigure}
     \hfill
     \begin{subfigure}[b]{0.49\textwidth}
         \centering
         \includegraphics{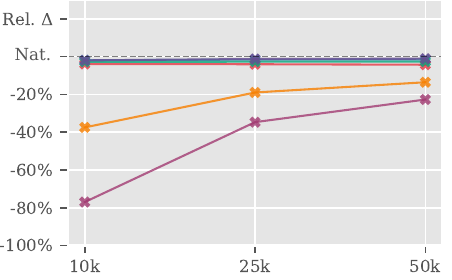}
         \caption{}
         \label{fig:uconv_loss_vs_vocab}
     \end{subfigure}
        \caption{Fertility vs. Token Collapse for monolingual models using \texttt{UConv}: \textbf{(a)} relative change in fertility (lower is better) for \mouconv compared to \moortho; \textbf{(b)} relative change in the vocabulary size (higher is better) of \moortho when romanizing each subword in its vocabulary using \texttt{UConv} (i.e., subwords get merged due to the same romanization).}
        \label{fig:uconv_efficiency_loss}
\end{figure*}

\begin{figure*}[h]
     \centering
     \begin{subfigure}[b]{0.49\textwidth}
         \centering
         \includegraphics{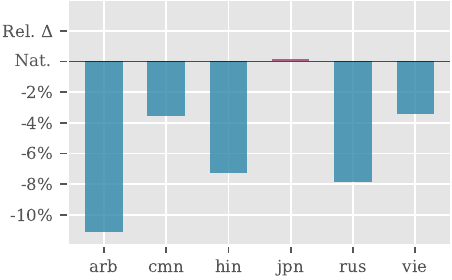}
         \caption{}
         \label{fig:multi_fertility_vs_vocab}
     \end{subfigure}
     \hfill
     \begin{subfigure}[b]{0.49\textwidth}
         \centering
         \includegraphics{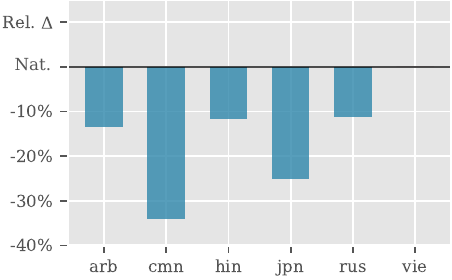}
         \caption{}
         \label{fig:multi_loss_vs_vocab}
     \end{subfigure}
        \caption{Fertility vs. token merging for multilingual models: \textbf{(a)} relative change in fertility (lower is better) for \mouroman compared to \moortho; \textbf{(b)} relative change in the vocabulary size (higher is better) of \moortho when romanizing each subword in its vocabulary (i.e., subwords get merged due to the same romanization).}
        \label{fig:multi_efficiency_loss}
\end{figure*}

\begin{table*}[h]
\setlength{\tabcolsep}{5.3pt}
\scriptsize
\centering
\begin{tabular}{lccccccccccc}
\toprule
 & \tokhp & \vocab & arb\_Arab & cmn\_Hani & hin\_Deva & jpn\_Jpan & rus\_Cyrl & vie\_Latn & \sequence & \token & Avg. \\
\midrule
\tmoortho & no-ws & 50k & 80.3$_{\pm0.6}$ & 78.4$_{\pm0.7}$ & 81.0$_{\pm1.6}$ & 74.9$_{\pm1.2}$ & 81.8$_{\pm0.6}$ & 82.0$_{\pm0.7}$ & 82.2$_{\pm1.0}$ & 76.6$_{\pm1.0}$ & 79.7$_{\pm1.0}$ \\
\tmoortho & ws & 50k & 79.8$_{\pm0.9}$ & 78.4$_{\pm1.0}$ & 81.3$_{\pm1.0}$ & 74.6$_{\pm1.7}$ & 81.8$_{\pm0.8}$ & 82.4$_{\pm0.5}$ & 82.0$_{\pm1.0}$ & 76.9$_{\pm1.1}$ & 79.7$_{\pm1.1}$ \\
\tmoortho & no-ws & 25k & 80.0$_{\pm1.2}$ & 78.1$_{\pm0.8}$ & 80.2$_{\pm0.7}$ & 75.4$_{\pm0.8}$ & 81.5$_{\pm0.9}$ & 82.0$_{\pm0.9}$ & 81.8$_{\pm0.9}$ & 76.8$_{\pm0.9}$ & 79.6$_{\pm0.9}$ \\
\tmoortho & ws & 25k & 79.8$_{\pm1.3}$ & 78.1$_{\pm1.5}$ & 80.8$_{\pm0.8}$ & 74.9$_{\pm0.7}$ & 81.7$_{\pm0.9}$ & 82.2$_{\pm0.7}$ & 81.8$_{\pm1.1}$ & 76.8$_{\pm1.0}$ & 79.6$_{\pm1.0}$ \\
\tmoortho & no-ws & 10k & 79.4$_{\pm0.8}$ & 79.1$_{\pm0.9}$ & 80.8$_{\pm1.5}$ & 75.6$_{\pm1.1}$ & 80.8$_{\pm1.2}$ & 82.0$_{\pm0.9}$ & 81.8$_{\pm1.1}$ & 76.9$_{\pm1.1}$ & 79.6$_{\pm1.1}$ \\
\tmoortho & ws & 10k & 79.3$_{\pm1.2}$ & 79.0$_{\pm0.9}$ & 80.3$_{\pm0.7}$ & 75.7$_{\pm0.9}$ & 81.0$_{\pm1.6}$ & 82.1$_{\pm0.9}$ & 81.7$_{\pm1.1}$ & 76.9$_{\pm1.1}$ & 79.6$_{\pm1.1}$ \\
\tmouroman & no-ws & 50k & 80.7$_{\pm0.8}$ & 74.1$_{\pm1.0}$ & 80.7$_{\pm0.8}$ & 71.7$_{\pm1.1}$ & 81.6$_{\pm0.6}$ & 81.9$_{\pm0.7}$ & 81.5$_{\pm0.8}$ & 74.7$_{\pm0.8}$ & 78.4$_{\pm0.8}$ \\
\tmouroman & ws & 50k & 80.1$_{\pm1.1}$ & 74.0$_{\pm0.8}$ & 80.8$_{\pm0.8}$ & 72.1$_{\pm0.8}$ & 81.7$_{\pm0.9}$ & 81.3$_{\pm0.8}$ & 81.1$_{\pm0.9}$ & 74.9$_{\pm0.9}$ & 78.3$_{\pm0.9}$ \\
\tmouroman & no-ws & 25k & 79.6$_{\pm0.6}$ & 73.6$_{\pm0.9}$ & 80.4$_{\pm1.0}$ & 71.1$_{\pm0.6}$ & 81.5$_{\pm1.1}$ & 80.9$_{\pm0.5}$ & 80.5$_{\pm0.8}$ & 74.6$_{\pm0.8}$ & 77.8$_{\pm0.8}$ \\
\tmouroman & ws & 25k & 79.5$_{\pm0.7}$ & 74.0$_{\pm1.5}$ & 80.9$_{\pm0.9}$ & 71.4$_{\pm0.8}$ & 81.5$_{\pm0.6}$ & 81.1$_{\pm1.1}$ & 80.7$_{\pm1.0}$ & 74.9$_{\pm1.0}$ & 78.1$_{\pm1.0}$ \\
\tmouroman & no-ws & 10k & 79.7$_{\pm1.1}$ & 74.7$_{\pm1.0}$ & 80.1$_{\pm0.8}$ & 71.8$_{\pm1.2}$ & 80.5$_{\pm1.1}$ & 80.3$_{\pm1.2}$ & 80.6$_{\pm1.1}$ & 74.5$_{\pm1.1}$ & 77.8$_{\pm1.1}$ \\
\tmouroman & ws & 10k & 79.3$_{\pm1.4}$ & 73.9$_{\pm1.1}$ & 80.8$_{\pm0.9}$ & 71.8$_{\pm0.8}$ & 80.4$_{\pm1.1}$ & 81.0$_{\pm0.7}$ & 80.5$_{\pm1.0}$ & 74.7$_{\pm1.0}$ & 77.9$_{\pm1.0}$ \\
\bottomrule
\end{tabular}
\caption{Main results averaged across our 5 evaluation tasks for monolingual and multilingual romanized models and different vocab sizes. We ablate different settings for the tokenizer training flag \textit{Split by Whitespace}: \texttt{ws} refers to \textit{Split by Whitespace} set to \textit{true}, \texttt{no-ws} refers to \textit{Split by Whitespace} set to \textit{false}.}
\label{tab:results_vocab_vs_results}
\end{table*}

\begin{table*}[h]
\setlength{\tabcolsep}{6.2pt}
\small
\centering
\begin{tabular}{lccccccccc}
\toprule
 & \tokhp & \vocab & arb\_Arab & cmn\_Hani & hin\_Deva & jpn\_Jpan & rus\_Cyrl & vie\_Latn & Avg. \\
\midrule
\tmoortho & no-ws & 50k & 86.2$_{\pm0.9}$ & 89.4$_{\pm1.0}$ & 85.9$_{\pm3.0}$ & 87.3$_{\pm2.0}$ & 86.8$_{\pm0.8}$ & 87.2$_{\pm0.9}$ & 87.1$_{\pm1.7}$ \\
\tmoortho & ws & 50k & 83.0$_{\pm1.2}$ & 88.3$_{\pm1.3}$ & 84.9$_{\pm1.7}$ & 86.6$_{\pm3.3}$ & 85.8$_{\pm1.3}$ & 88.8$_{\pm0.7}$ & 86.2$_{\pm1.8}$ \\
\tmoortho & no-ws & 25k & 85.0$_{\pm2.6}$ & 86.2$_{\pm1.5}$ & 82.3$_{\pm0.9}$ & 88.0$_{\pm1.4}$ & 85.6$_{\pm1.6}$ & 88.7$_{\pm1.7}$ & 86.0$_{\pm1.7}$ \\
\tmoortho & ws & 25k & 84.7$_{\pm2.5}$ & 85.4$_{\pm3.1}$ & 84.1$_{\pm1.4}$ & 87.9$_{\pm1.2}$ & 87.1$_{\pm1.3}$ & 89.8$_{\pm1.3}$ & 86.5$_{\pm1.9}$ \\
\tmoortho & no-ws & 10k & 84.5$_{\pm1.5}$ & 88.0$_{\pm1.0}$ & 87.3$_{\pm3.0}$ & 86.3$_{\pm1.8}$ & 84.5$_{\pm2.0}$ & 89.3$_{\pm1.5}$ & 86.7$_{\pm1.9}$ \\
\tmoortho & ws & 10k & 83.0$_{\pm2.5}$ & 88.0$_{\pm1.9}$ & 85.3$_{\pm0.7}$ & 87.8$_{\pm1.4}$ & 85.9$_{\pm3.5}$ & 89.4$_{\pm1.6}$ & 86.6$_{\pm2.1}$ \\
\tmouroman & no-ws & 50k & 88.2$_{\pm1.3}$ & 84.5$_{\pm1.7}$ & 84.5$_{\pm1.1}$ & 84.7$_{\pm2.0}$ & 85.8$_{\pm0.7}$ & 89.6$_{\pm1.1}$ & 86.2$_{\pm1.4}$ \\
\tmouroman & ws & 50k & 85.3$_{\pm2.2}$ & 84.7$_{\pm1.4}$ & 84.6$_{\pm1.2}$ & 84.7$_{\pm1.3}$ & 85.6$_{\pm1.5}$ & 87.9$_{\pm1.1}$ & 85.5$_{\pm1.5}$ \\
\tmouroman & no-ws & 25k & 85.0$_{\pm0.7}$ & 83.5$_{\pm1.7}$ & 82.7$_{\pm2.0}$ & 81.8$_{\pm0.4}$ & 86.6$_{\pm1.9}$ & 87.2$_{\pm0.7}$ & 84.5$_{\pm1.4}$ \\
\tmouroman & ws & 25k & 84.0$_{\pm0.7}$ & 83.2$_{\pm2.9}$ & 85.8$_{\pm1.7}$ & 83.1$_{\pm1.0}$ & 86.3$_{\pm0.6}$ & 87.8$_{\pm2.2}$ & 85.0$_{\pm1.7}$ \\
\tmouroman & no-ws & 10k & 86.8$_{\pm2.0}$ & 84.1$_{\pm2.0}$ & 83.6$_{\pm1.0}$ & 84.3$_{\pm2.2}$ & 84.3$_{\pm2.1}$ & 84.6$_{\pm2.1}$ & 84.6$_{\pm2.0}$ \\
\tmouroman & ws & 10k & 83.7$_{\pm2.9}$ & 84.6$_{\pm1.7}$ & 85.2$_{\pm1.7}$ & 83.6$_{\pm1.0}$ & 83.8$_{\pm2.1}$ & 87.3$_{\pm0.8}$ & 84.7$_{\pm1.9}$ \\
\bottomrule
\end{tabular}
\caption{SIB200 results for monolingual and multilingual romanized models and different vocab sizes. We ablate different settings for the tokenizer training flag \textit{Split by Whitespace}: \texttt{ws} refers to \textit{Split by Whitespace} set to \textit{true}, \texttt{no-ws} refers to \textit{Split by Whitespace} set to \textit{false}.}
\label{tab:results_sib200_vocab}
\end{table*}

\begin{table*}[h]
\setlength{\tabcolsep}{6.2pt}
\small
\centering
\begin{tabular}{lccccccccc}
\toprule
 & \tokhp & \vocab & arb\_Arab & cmn\_Hani & hin\_Deva & jpn\_Jpan & rus\_Cyrl & vie\_Latn & Avg. \\
\midrule
\tmoortho & no-ws & 50k & 89.7$_{\pm0.1}$ & 75.4$_{\pm0.2}$ & 90.1$_{\pm0.5}$ & 65.9$_{\pm0.2}$ & 87.3$_{\pm0.1}$ & 90.5$_{\pm0.1}$ & 83.2$_{\pm0.2}$ \\
\tmoortho & ws & 50k & 89.9$_{\pm0.1}$ & 75.3$_{\pm0.5}$ & 91.3$_{\pm0.5}$ & 65.9$_{\pm0.3}$ & 87.6$_{\pm0.1}$ & 91.0$_{\pm0.2}$ & 83.5$_{\pm0.3}$ \\
\tmoortho & no-ws & 25k & 89.6$_{\pm0.1}$ & 76.6$_{\pm0.4}$ & 90.5$_{\pm0.6}$ & 66.3$_{\pm0.2}$ & 87.2$_{\pm0.1}$ & 90.3$_{\pm0.2}$ & 83.4$_{\pm0.3}$ \\
\tmoortho & ws & 25k & 89.5$_{\pm0.3}$ & 76.6$_{\pm0.2}$ & 90.6$_{\pm0.6}$ & 66.3$_{\pm0.4}$ & 87.6$_{\pm0.2}$ & 90.8$_{\pm0.0}$ & 83.5$_{\pm0.3}$ \\
\tmoortho & no-ws & 10k & 89.0$_{\pm0.2}$ & 77.6$_{\pm0.4}$ & 90.2$_{\pm0.5}$ & 67.7$_{\pm0.2}$ & 87.1$_{\pm0.2}$ & 90.3$_{\pm0.2}$ & 83.6$_{\pm0.3}$ \\
\tmoortho & ws & 10k & 89.5$_{\pm0.2}$ & 77.6$_{\pm0.4}$ & 89.9$_{\pm0.7}$ & 66.9$_{\pm0.3}$ & 86.9$_{\pm0.1}$ & 90.7$_{\pm0.2}$ & 83.6$_{\pm0.4}$ \\
\tmouroman & no-ws & 50k & 89.6$_{\pm0.2}$ & 68.1$_{\pm0.4}$ & 89.5$_{\pm0.7}$ & 59.7$_{\pm0.1}$ & 87.1$_{\pm0.3}$ & 89.5$_{\pm0.3}$ & 80.5$_{\pm0.4}$ \\
\tmouroman & ws & 50k & 89.7$_{\pm0.2}$ & 67.9$_{\pm0.2}$ & 90.4$_{\pm0.4}$ & 59.3$_{\pm0.3}$ & 87.4$_{\pm0.1}$ & 90.2$_{\pm0.1}$ & 80.8$_{\pm0.2}$ \\
\tmouroman & no-ws & 25k & 89.2$_{\pm0.1}$ & 67.9$_{\pm0.4}$ & 90.7$_{\pm0.5}$ & 59.2$_{\pm0.4}$ & 87.0$_{\pm0.1}$ & 89.4$_{\pm0.1}$ & 80.5$_{\pm0.3}$ \\
\tmouroman & ws & 25k & 89.6$_{\pm0.2}$ & 68.9$_{\pm0.3}$ & 90.2$_{\pm0.5}$ & 59.6$_{\pm0.2}$ & 87.1$_{\pm0.2}$ & 90.1$_{\pm0.3}$ & 80.9$_{\pm0.3}$ \\
\tmouroman & no-ws & 10k & 88.9$_{\pm0.2}$ & 69.2$_{\pm0.1}$ & 89.9$_{\pm0.7}$ & 60.0$_{\pm0.6}$ & 86.6$_{\pm0.2}$ & 89.5$_{\pm0.3}$ & 80.7$_{\pm0.4}$ \\
\tmouroman & ws & 10k & 89.1$_{\pm0.3}$ & 68.7$_{\pm0.3}$ & 90.7$_{\pm0.5}$ & 60.5$_{\pm0.5}$ & 87.0$_{\pm0.2}$ & 90.2$_{\pm0.1}$ & 81.0$_{\pm0.3}$ \\
\bottomrule
\end{tabular}
\caption{Wikiann results for monolingual and multilingual romanized models and different vocab sizes. We ablate different settings for the tokenizer training flag \textit{Split by Whitespace}: \texttt{ws} refers to \textit{Split by Whitespace} set to \textit{true}, \texttt{no-ws} refers to \textit{Split by Whitespace} set to \textit{false}.}
\label{tab:results_wikiann_vocab}
\end{table*}

\begin{table*}[h]
\setlength{\tabcolsep}{8.7pt}
\small
\centering
\begin{tabular}{lcccccccc}
\toprule
 & \tokhp & \vocab & arb\_Arab & cmn\_Hani & hin\_Deva & rus\_Cyrl & vie\_Latn & Avg. \\
\midrule
\tmoortho & no-ws & 50k & 74.6$_{\pm0.6}$ & 74.8$_{\pm0.7}$ & 72.0$_{\pm1.5}$ & 75.7$_{\pm1.0}$ & 76.7$_{\pm1.1}$ & 74.8$_{\pm1.1}$ \\
\tmoortho & ws & 50k & 74.2$_{\pm1.3}$ & 75.3$_{\pm1.0}$ & 72.8$_{\pm1.0}$ & 75.6$_{\pm0.9}$ & 76.4$_{\pm0.7}$ & 74.9$_{\pm1.0}$ \\
\tmoortho & no-ws & 25k & 74.1$_{\pm0.5}$ & 74.6$_{\pm0.9}$ & 71.6$_{\pm0.9}$ & 75.9$_{\pm0.4}$ & 76.2$_{\pm0.7}$ & 74.5$_{\pm0.7}$ \\
\tmoortho & ws & 25k & 74.0$_{\pm0.4}$ & 74.7$_{\pm1.0}$ & 71.7$_{\pm0.8}$ & 74.6$_{\pm1.3}$ & 75.4$_{\pm0.6}$ & 74.1$_{\pm0.9}$ \\
\tmoortho & no-ws & 10k & 73.8$_{\pm0.7}$ & 74.0$_{\pm1.4}$ & 70.9$_{\pm0.8}$ & 74.0$_{\pm1.3}$ & 75.9$_{\pm1.1}$ & 73.7$_{\pm1.1}$ \\
\tmoortho & ws & 10k & 73.8$_{\pm0.7}$ & 73.9$_{\pm0.5}$ & 71.1$_{\pm1.0}$ & 73.9$_{\pm0.9}$ & 75.0$_{\pm0.6}$ & 73.5$_{\pm0.8}$ \\
\tmouroman & no-ws & 50k & 74.2$_{\pm0.7}$ & 71.2$_{\pm1.2}$ & 73.3$_{\pm0.6}$ & 75.9$_{\pm0.8}$ & 76.3$_{\pm0.9}$ & 74.2$_{\pm0.9}$ \\
\tmouroman & ws & 50k & 74.9$_{\pm0.9}$ & 70.7$_{\pm0.8}$ & 72.9$_{\pm0.7}$ & 75.4$_{\pm1.1}$ & 74.5$_{\pm0.7}$ & 73.7$_{\pm0.9}$ \\
\tmouroman & no-ws & 25k & 73.7$_{\pm0.9}$ & 70.4$_{\pm0.9}$ & 72.1$_{\pm0.8}$ & 75.1$_{\pm0.9}$ & 75.6$_{\pm0.5}$ & 73.4$_{\pm0.8}$ \\
\tmouroman & ws & 25k & 73.8$_{\pm1.0}$ & 70.7$_{\pm1.3}$ & 72.3$_{\pm0.6}$ & 74.5$_{\pm0.8}$ & 74.2$_{\pm0.9}$ & 73.1$_{\pm0.9}$ \\
\tmouroman & no-ws & 10k & 73.2$_{\pm1.1}$ & 71.0$_{\pm0.5}$ & 72.0$_{\pm1.0}$ & 74.5$_{\pm0.8}$ & 75.6$_{\pm0.9}$ & 73.3$_{\pm0.9}$ \\
\tmouroman & ws & 10k & 73.8$_{\pm0.6}$ & 70.8$_{\pm1.0}$ & 71.8$_{\pm0.7}$ & 73.4$_{\pm1.1}$ & 73.9$_{\pm0.6}$ & 72.8$_{\pm0.8}$ \\
\bottomrule
\end{tabular}
\caption{XNLI results for monolingual and multilingual romanized models and different vocab sizes. We ablate different settings for the tokenizer training flag \textit{Split by Whitespace}: \texttt{ws} refers to \textit{Split by Whitespace} set to \textit{true}, \texttt{no-ws} refers to \textit{Split by Whitespace} set to \textit{false}.}
\label{tab:results_xnli_vocab}
\end{table*}

\begin{table*}[h]
\setlength{\tabcolsep}{6.2pt}
\small
\centering
\begin{tabular}{lccccccccc}
\toprule
 & \tokhp & \vocab & arb\_Arab & cmn\_Hani & hin\_Deva & jpn\_Jpan & rus\_Cyrl & vie\_Latn & Avg. \\
\midrule
\tmoortho & no-ws & 50k & 79.8$_{\pm0.5}$ & 83.6$_{\pm0.5}$ & 85.1$_{\pm0.5}$ & 83.2$_{\pm0.5}$ & 84.8$_{\pm0.5}$ & 85.0$_{\pm0.2}$ & 83.6$_{\pm0.5}$ \\
\tmoortho & ws & 50k & 79.9$_{\pm0.5}$ & 83.3$_{\pm1.3}$ & 85.1$_{\pm0.5}$ & 83.3$_{\pm0.5}$ & 85.3$_{\pm0.6}$ & 85.3$_{\pm0.3}$ & 83.7$_{\pm0.7}$ \\
\tmoortho & no-ws & 25k & 79.6$_{\pm0.5}$ & 83.6$_{\pm0.3}$ & 85.3$_{\pm0.4}$ & 83.3$_{\pm0.2}$ & 84.8$_{\pm1.1}$ & 85.1$_{\pm0.2}$ & 83.6$_{\pm0.5}$ \\
\tmoortho & ws & 25k & 79.2$_{\pm1.4}$ & 84.2$_{\pm0.4}$ & 84.9$_{\pm0.3}$ & 82.7$_{\pm0.4}$ & 85.3$_{\pm0.7}$ & 85.2$_{\pm0.6}$ & 83.6$_{\pm0.7}$ \\
\tmoortho & no-ws & 10k & 79.1$_{\pm0.5}$ & 84.8$_{\pm0.4}$ & 84.1$_{\pm0.5}$ & 83.6$_{\pm0.5}$ & 84.9$_{\pm0.8}$ & 84.8$_{\pm0.1}$ & 83.5$_{\pm0.5}$ \\
\tmoortho & ws & 10k & 79.1$_{\pm0.7}$ & 84.7$_{\pm0.1}$ & 83.7$_{\pm0.5}$ & 83.1$_{\pm0.6}$ & 85.2$_{\pm0.3}$ & 85.4$_{\pm0.3}$ & 83.5$_{\pm0.5}$ \\
\tmouroman & no-ws & 50k & 79.4$_{\pm0.6}$ & 81.4$_{\pm0.4}$ & 84.5$_{\pm0.9}$ & 82.1$_{\pm0.3}$ & 85.3$_{\pm0.6}$ & 84.2$_{\pm0.6}$ & 82.8$_{\pm0.6}$ \\
\tmouroman & ws & 50k & 79.3$_{\pm0.3}$ & 81.5$_{\pm0.6}$ & 84.6$_{\pm0.6}$ & 82.4$_{\pm0.4}$ & 85.5$_{\pm0.6}$ & 84.4$_{\pm0.8}$ & 83.0$_{\pm0.6}$ \\
\tmouroman & no-ws & 25k & 79.0$_{\pm0.7}$ & 81.1$_{\pm0.6}$ & 84.6$_{\pm0.3}$ & 82.0$_{\pm0.4}$ & 84.7$_{\pm0.8}$ & 84.0$_{\pm0.5}$ & 82.6$_{\pm0.6}$ \\
\tmouroman & ws & 25k & 79.0$_{\pm0.4}$ & 81.4$_{\pm0.8}$ & 84.3$_{\pm0.2}$ & 81.9$_{\pm0.3}$ & 84.9$_{\pm0.8}$ & 84.4$_{\pm0.4}$ & 82.7$_{\pm0.5}$ \\
\tmouroman & no-ws & 10k & 78.8$_{\pm0.6}$ & 82.5$_{\pm0.6}$ & 84.0$_{\pm0.8}$ & 82.5$_{\pm0.2}$ & 84.2$_{\pm0.9}$ & 83.6$_{\pm1.4}$ & 82.6$_{\pm0.8}$ \\
\tmouroman & ws & 10k & 78.9$_{\pm0.5}$ & 81.2$_{\pm0.7}$ & 84.8$_{\pm0.6}$ & 82.3$_{\pm0.3}$ & 84.3$_{\pm0.6}$ & 84.6$_{\pm0.5}$ & 82.7$_{\pm0.5}$ \\
\bottomrule
\end{tabular}
\caption{MASSIVE Intent Classification results for monolingual and multilingual romanized models and different vocab sizes. We ablate different settings for the tokenizer training flag \textit{Split by Whitespace}: \texttt{ws} refers to \textit{Split by Whitespace} set to \textit{true}, \texttt{no-ws} refers to \textit{Split by Whitespace} set to \textit{false}.}
\label{tab:results_massive_intent_vocab}
\end{table*}

\begin{table*}[h]
\setlength{\tabcolsep}{6.2pt}
\small
\centering
\begin{tabular}{lccccccccc}
\toprule
 & \tokhp & \vocab & arb\_Arab & cmn\_Hani & hin\_Deva & jpn\_Jpan & rus\_Cyrl & vie\_Latn & Avg. \\
\midrule
\tmoortho & no-ws & 50k & 71.3$_{\pm0.6}$ & 68.6$_{\pm0.9}$ & 71.9$_{\pm0.4}$ & 63.3$_{\pm1.4}$ & 74.6$_{\pm0.1}$ & 70.5$_{\pm0.6}$ & 70.0$_{\pm0.8}$ \\
\tmoortho & ws & 50k & 72.1$_{\pm0.6}$ & 69.5$_{\pm0.6}$ & 72.4$_{\pm0.6}$ & 62.6$_{\pm0.8}$ & 74.7$_{\pm0.3}$ & 70.5$_{\pm0.2}$ & 70.3$_{\pm0.6}$ \\
\tmoortho & no-ws & 25k & 71.9$_{\pm0.6}$ & 69.6$_{\pm0.5}$ & 71.4$_{\pm0.3}$ & 64.0$_{\pm0.4}$ & 74.3$_{\pm0.6}$ & 69.8$_{\pm0.5}$ & 70.2$_{\pm0.5}$ \\
\tmoortho & ws & 25k & 71.6$_{\pm0.2}$ & 69.5$_{\pm0.5}$ & 72.5$_{\pm0.4}$ & 62.8$_{\pm0.3}$ & 74.2$_{\pm0.5}$ & 69.9$_{\pm0.4}$ & 70.1$_{\pm0.4}$ \\
\tmoortho & no-ws & 10k & 70.6$_{\pm0.5}$ & 71.1$_{\pm1.0}$ & 71.4$_{\pm0.9}$ & 65.0$_{\pm1.0}$ & 73.6$_{\pm0.8}$ & 69.8$_{\pm0.4}$ & 70.2$_{\pm0.8}$ \\
\tmoortho & ws & 10k & 70.9$_{\pm0.7}$ & 70.8$_{\pm0.8}$ & 71.7$_{\pm0.5}$ & 65.0$_{\pm0.8}$ & 73.4$_{\pm0.8}$ & 70.0$_{\pm1.0}$ & 70.3$_{\pm0.8}$ \\
\tmouroman & no-ws & 50k & 72.2$_{\pm0.8}$ & 65.2$_{\pm0.6}$ & 71.6$_{\pm0.5}$ & 60.4$_{\pm0.8}$ & 74.2$_{\pm0.4}$ & 69.7$_{\pm0.6}$ & 68.9$_{\pm0.6}$ \\
\tmouroman & ws & 50k & 71.5$_{\pm0.5}$ & 65.2$_{\pm0.5}$ & 71.5$_{\pm0.6}$ & 62.0$_{\pm0.8}$ & 74.4$_{\pm0.6}$ & 69.6$_{\pm0.9}$ & 69.0$_{\pm0.7}$ \\
\tmouroman & no-ws & 25k & 71.0$_{\pm0.3}$ & 65.2$_{\pm0.4}$ & 71.6$_{\pm0.6}$ & 61.2$_{\pm1.0}$ & 74.2$_{\pm0.5}$ & 68.5$_{\pm0.6}$ & 68.6$_{\pm0.6}$ \\
\tmouroman & ws & 25k & 71.1$_{\pm0.9}$ & 65.6$_{\pm0.9}$ & 71.8$_{\pm0.6}$ & 60.9$_{\pm1.2}$ & 74.6$_{\pm0.2}$ & 69.1$_{\pm0.7}$ & 68.9$_{\pm0.8}$ \\
\tmouroman & no-ws & 10k & 71.0$_{\pm0.8}$ & 66.5$_{\pm0.8}$ & 70.8$_{\pm0.4}$ & 60.3$_{\pm0.7}$ & 72.9$_{\pm0.6}$ & 68.4$_{\pm0.8}$ & 68.3$_{\pm0.7}$ \\
\tmouroman & ws & 10k & 71.0$_{\pm0.4}$ & 64.0$_{\pm1.0}$ & 71.4$_{\pm0.4}$ & 61.0$_{\pm1.0}$ & 73.5$_{\pm0.6}$ & 69.3$_{\pm1.0}$ & 68.4$_{\pm0.8}$ \\
\bottomrule
\end{tabular}
\caption{MASSIVE Slot Filling results for monolingual and multilingual romanized models and different vocab sizes. We ablate different settings for the tokenizer training flag \textit{Split by Whitespace}: \texttt{ws} refers to \textit{Split by Whitespace} set to \textit{true}, \texttt{no-ws} refers to \textit{Split by Whitespace} set to \textit{false}.}
\label{tab:results_massive_slot_vocab}
\end{table*}



\end{document}